\pdfoutput=1
\documentclass{article}

\PassOptionsToPackage{numbers}{natbib}
\usepackage[preprint]{neurips_2023}
\usepackage{multirow}
\usepackage{graphicx}
\usepackage{makecell}
\usepackage{subcaption}

\usepackage{bm}
\usepackage[utf8]{inputenc} 
\usepackage[T1]{fontenc}    
\usepackage{hyperref}       
\usepackage{url}            
\usepackage{booktabs}       
\usepackage{amsfonts}       
\usepackage{nicefrac}       
\usepackage{microtype}      
\usepackage{xcolor}         

\usepackage{tcolorbox}

\title{Normalization Enhances Generalization in Visual Reinforcement Learning}

\author{
Lu Li$^1$\thanks{Equal contribution.}
\quad
Jiafei Lyu$^1$\footnotemark[1]
\quad
Guozheng Ma$^1$
\quad
Zilin Wang$^1$ \\
\vspace{0.2cm}
\textbf{
Zhenjie Yang$^2$ 
\quad
Xiu Li$^{1 \dagger}$
\quad
Zhiheng Li$^1$\thanks{Corresponding authors.}
}\\
\vspace{0.2cm}
$^1$Tsinghua University
\quad
$^2$SenseTime Research \\
\texttt{\{lilu21, lvjf20\}@mails.tsinghua.edu.cn, zhhli@tsinghua.edu.cn}
}

\begin{document}

\maketitle

\begin{abstract}
Recent advances in visual reinforcement learning (RL) have led to impressive success in handling complex tasks. However, these methods have demonstrated limited generalization capability to visual disturbances, which poses a significant challenge for their real-world application and adaptability.
Though normalization techniques have demonstrated huge success in supervised and unsupervised learning, their applications in visual RL are still scarce.
In this paper, we explore the potential benefits of integrating normalization into visual RL methods with respect to generalization performance. We find that, perhaps surprisingly, incorporating suitable normalization techniques is sufficient to enhance the generalization capabilities, without any additional special design. We utilize the combination of two normalization techniques, CrossNorm and SelfNorm, for generalizable visual RL. Extensive experiments are conducted on DMControl Generalization Benchmark and CARLA to validate the effectiveness of our method. We show that our method significantly improves generalization capability while only marginally affecting sample efficiency. In particular, when integrated with DrQ-v2, our method enhances the test performance of DrQ-v2 on CARLA across various scenarios, from 14\% of the training performance to  \textbf{97\%}. Our project page: \url{https://sites.google.com/view/norm-generalization-vrl/home}

\end{abstract}
\section{Introduction}

Visual reinforcement learning (RL), which leverages high-dimensional visual observations as inputs, has shown  potential in a wide range of tasks, such as playing video games~\cite{mnih2013playing,vinyals2019grandmaster} and robotic manipulation~\cite{levine2016end}. However, generalization remains a major challenge for visual RL methods. Even slight alterations, such as color or background changes, can result in considerable performance degradation in the testing environment, which in turn limits the real-world utility of these algorithms. In light of these challenges, it is essential to develop techniques that can improve the generalization capabilities of visual RL algorithms.

Existing literature mainly enhances the generalization capability of visual RL via data augmentation~\cite{svea,tlda,pmlr-v139-fan21c,wang2020improving} and domain randomization~\cite{tobin2017domain,pinto2017asymmetric,peng2018sim}, aiming at learning policies invariant to the changes in the observations. However, recent studies \cite{yarats2021image,rad} show that certain data augmentation techniques may lead to a decrease in sample efficiency and even cause divergence. Other recent works improve the generalization performance by leveraging pre-trained image encoder \cite{Yuan2022PreTrainedIE} or segmenting important pixels from the test environment \cite{Bertoin2022LookWY}, \textit{etc.} Unfortunately, most of them rely on knowledge or data from outer sources, \textit{e.g.}, ImageNet \cite{deng2009imagenet}.
We deem that an ideal method for zero-shot generalization should be able to achieve robust performance without relying on any out-of-domain data or prior knowledge of the target domain, and should be able to adapt effectively to a wide variety of environments and tasks.

Normalization techniques have achieved huge success in computer vision~\cite{ulyanov2017instance,Wu2018GroupN,tang2021crossnorm} and natural language processing~\cite{ba2016layer,Xiong2020OnLN,Wang2019MultipassageBA}. Numerous normalization-related methods are proposed to improve the generalization capabilities of deep neural networks \cite{Sanyal2019StableRN,Seo2019LearningTO,fan2021adversarially,huang2023normalization}. Despite their popularity, normalization techniques seem to have not received sufficient attention in deep RL community. Though previous studies have investigated the effectiveness of normalization methods, \textit{e.g.}, layer normalization \cite{Hiraoka2021DropoutQF,Parisotto2019StabilizingTF} and spectral normalization \cite{miyato2018spectral,bjorck2021towards,Mehta2022EffectsOS,gogianu2021spectral}, in deep RL algorithms, to the best of our knowledge, it is still unclear whether normalization can aid generalization in visual RL. Building upon these insights, we would like to ask the following question: 

\begin{center}
\textit{Can we develop a visual RL agent that employs normalization techniques and does not rely on prior knowledge and out-of-domain data, enabling it to generalize more effectively to unseen scenarios? }
\end{center}
This inquiry drives our exploration of CrossNorm and SelfNorm \cite{tang2021crossnorm}, two normalization methods that have been proven to enhance generalization in computer vision tasks under distribution shifts. 
Since visual RL algorithms always rely on the encoder to output representations for policy learning and action execution, we need to ensure that the learned representation can generalize to unseen scenarios. To fulfill that, we propose to modify the encoder structure of the base visual RL algorithm by incorporating CrossNorm and SelfNorm for the downstream tasks. Our proposed normalization module is plug-and-play, and can be combined with any existing visual RL algorithms. 

We evaluate the performance of our method on DeepMind Control Generalization Benchmark ~\cite{hansen2021softda}, a benchmark designed for evaluating generalization capabilities in robotic control tasks, and CARLA~\cite{Dosovitskiy17}, a realistic autonomous driving simulator. Extensive experimental results demonstrate that when combined with DrQ~\cite{yarats2021image} and DrQ-v2~\cite{yarats2021drqv2}, our proposed normalization module significantly improves their generalization capabilities without requiring any task-specific modifications or prior knowledge. 
Furthermore, our proposed module demonstrates compatibility and synergy with other generalization algorithms in visual RL (\textit{e.g.}, SVEA \cite{svea}), thereby further enhance their  generalization. This indicates the flexibility of our proposed module and its potential to be a valuable addition to the toolset for improving generalization in visual RL tasks. We believe this work offers another chance that allows visual RL algorithms to exhibit greater adaptability and robustness across diverse and dynamic environments. We aspire to propel the field of visual RL forward and broaden the scope of the potential applications of normalization techniques.

\section{Related Work}
\subsection{Generalization in Visual RL}

Over the past few years, considerable strides have been made towards narrowing the generalization gap in visual RL.
An elementary strategy for improving generalization is to employ regularization techniques, initially developed for supervised learning~\cite{regularization_matter}. These techniques include $\ell_2$ regularization~\cite{bn_rl}, entropy regularization~\cite{study_on_overfitting}, and dropout~\cite{Consistent_Dropout}.
Unfortunately, these conventional regularization techniques exhibit limited effectiveness in improving generalization of visual RL and, in some cases, they may even have a negative impact on sample efficiency~\cite{CoinRun, SNI_IB}.
As a result, recent studies have shifted their focus towards learning robust representations by leveraging bisimulation metrics~\citep{dbc, robust_bisimulation}, multi-view information bottleneck (MIB)~\citep{dribo}, pre-trained image encoder~\citep{Yuan2022PreTrainedIE}, \textit{etc}.
From an orthogonal perspective, data augmentation has demonstrated significant efficacy in enhancing generalization by leveraging prior knowledge as an inductive bias for the agent~\cite{rad, svea, tlda, DA_survey}. 
However, the effectiveness of data augmentation-based techniques is significantly constrained by their highly task-specific nature and the requirement for substantial expert knowledge~\cite{drac, Generalisation_survey, ma2023learning}.
One one hand, applying appropriate data augmentation techniques demands domain-specific knowledge, which limits their applicability to unfamiliar or novel environments. On the other hand, these techniques face challenges in generalizing to new domains due to their reliance on the alignment between augmentations and domain characteristics.
In this study, our objective is to explore the utilization of normalization techniques to enhance the generalizability of visual RL, without relying on specific prior knowledge of the shift characteristics between the train and test environments. We note that two recent studies~\cite{Generalisation_survey,DA_survey} present a comprehensive analysis of the generalization challenges in RL and the application of data augmentation in visual RL, which can be a nice reference.

\subsection{Normalization}

Normalization techniques play a crucial role in training deep neural networks \cite{Lubana2021BeyondBT,Salimans2016WeightNA,Wu2018GroupN}. They notably enhance optimization by normalizing input features, which is particularly advantageous for first-order optimization algorithms such as Stochastic Gradient Descent (SGD) \cite{Bottou2010LargeScaleML,Zinkevich2010ParallelizedSG}, known to excel in more isotropic landscapes~\cite{boyd2004convex}. Batch Normalization~\cite{ioffe2015batch,Bjorck2018UnderstandingBN,Santurkar2018HowDB} (BN) is a method that normalizes intermediate feature maps using statistics computed from mini-batch samples. This technique has been found to significantly aid in the training of deep networks. Drawing inspiration from the success of BN, a variety of normalization techniques have since been introduced to accommodate different learning scenarios,  \textit{e.g.}, layer normalization~\cite{ba2016layer,sun2020new,Zhang2019RootMS}, spectral normalization~\cite{miyato2018spectral,Lin2020WhySN}, \textit{etc}. 

Despite the huge success and wide applications of normalization techniques, they are not commonly employed in deep RL. This is largely attributed to the online learning nature of RL, which leads to a non-independent and identically distributed (non-i.i.d) input data distribution. Such distribution does not align with the requirements of many normalization techniques. ~\cite{bhatt2019crossnorm} shows that direct application of BN and LN proves to be ineffective for RL. Instead, it introduces cross-normalization, which computes mean feature subtraction using both on-policy and off-policy state-action pairs, leading to better sample efficiency.
Moreover, spectral normalization has been found to be effective in stabilizing the training process of RL \cite{bjorck2021towards,gogianu2021spectral}.

It is interesting to ask: since normalization techniques have shown benefits for generalization to new tasks in computer vision, then whether normalization techniques have the potential to enhance the generalization ability of the visual RL algorithms. To the best of our knowledge, none of the prior work explores this issue, and our goal in this work is to answer this question.

\section{Preliminary}
\subsection{Visual Reinforcement Learning}

We consider learning in a Partially Observable Markov Decision Processes (POMDPs) specified by the tuple $\mathcal{M}:\langle \mathcal{S},\mathcal{O},\mathcal{A},\mathcal{P},r,\gamma \rangle$, where $\mathcal{S}$ is the state space, $\mathcal{O}$ is the observation space, $\mathcal{A}$ is the action space, $\mathcal{P}(\cdot|s,a):\mathcal{S}\times\mathcal{A}\mapsto\mathbb{R}$ is the transition probability, $r(s,a):\mathcal{S}\times\mathcal{A}\mapsto\mathbb{R}$ is the scalar reward function, $\gamma\in[0,1)$ is the discount factor. In the context of generalization setting, we have a set of such POMDPs $M = \{\mathcal{M}_0,\mathcal{M}_1,\ldots,\mathcal{M}_n\}$ while our agent only has access to one fixed POMDP among them, denoted as $\mathcal{M}_0$. We aim to train an RL agent to learn a policy $\pi_\theta(\cdot|s)$ parameterized by the parameter $\theta$ in $\mathcal{M}_0$, with the objective of maximizing the expected cumulative return $J(\theta) = \mathbb{E}_{a_t\sim\pi_\theta(\cdot|s_t),s_t\sim\mathcal{P}}[\sum_{t=0}^T\gamma^t r(s_t,a_t)]$ across the entire set of POMDPs in a zero-shot manner, where $T$ is the horizon of the POMDP.

\subsection{CrossNorm and SelfNorm}
CrossNorm and SelfNorm~\cite{tang2021crossnorm} were initially introduced to improve generalization capabilities in the face of distribution shifts within computer vision tasks.
To broaden the training distribution, CrossNorm swaps the mean and standard deviation of channel $\mathcal{A}$, denoted as $\mu_\mathcal{A}$ and $\sigma_\mathcal{A}$ respectively, with the mean and standard deviation of channel $\mathcal{B}$, denoted as $\mu_\mathcal{B}$ and $\sigma_\mathcal{B}$ respectively. In other words, it exchanges $\mu$ and $\sigma$ values between channels $\mathcal{A}$ and $\mathcal{B}$, as shown in Equation \ref{eq:crossnorm}:
\begin{equation}\label{eq:crossnorm}
\mathcal{A'} = \sigma_\mathcal{B}\frac{\mathcal{A}-\mu_\mathcal{A}}{\sigma_\mathcal{A}}+\mu_\mathcal{B},
\qquad 
\mathcal{B'}=\sigma_\mathcal{A}\frac{\mathcal{B}-\mu_\mathcal{B}}{\sigma_\mathcal{B}}+\mu_\mathcal{A},
\end{equation}
While CrossNorm enlarges the training distribution, the motivation of SelfNorm is to bridge the train-test distribution gap. 
To achieve this, SelfNorm replaces $\mathcal{A}$ and $\mathcal{B}$ with recalibrated mean $\mu'_\mathcal{A}=f(\mu_\mathcal{A}, \sigma_\mathcal{A})\mu_\mathcal{A}$ and standard deviation $\sigma'_\mathcal{A}=g(\mu_\mathcal{A}, \sigma_\mathcal{A})\sigma_\mathcal{A}$, where $f$ and $g$ are the attention functions. The adjusted feature becomes as Equation \ref{eq:selfnorm}:
\begin{equation}\label{eq:selfnorm}
\hat\mathcal{A} = \sigma'_\mathcal{A}\frac{\mathcal{A}-\mu_\mathcal{A}}{\sigma_\mathcal{A}}+\mu'_\mathcal{A}.
\end{equation}
As $f$ and $g$ learn to scale $\mu_\mathcal{A}$ and $\sigma_\mathcal{A}$ based on their values, the method adapts to the specific characteristics of the data. While CrossNorm expands the data distribution, SelfNorm aims to emphasize the discriminative styles shared by both training and test distributions while de-emphasizing the insignificant styles.

\section{Method}

\subsection{Enhancing Generalization in Visual RL via Normalization}

\begin{figure}[t]
    \centering
    \centering
    \includegraphics[width=0.95\linewidth]{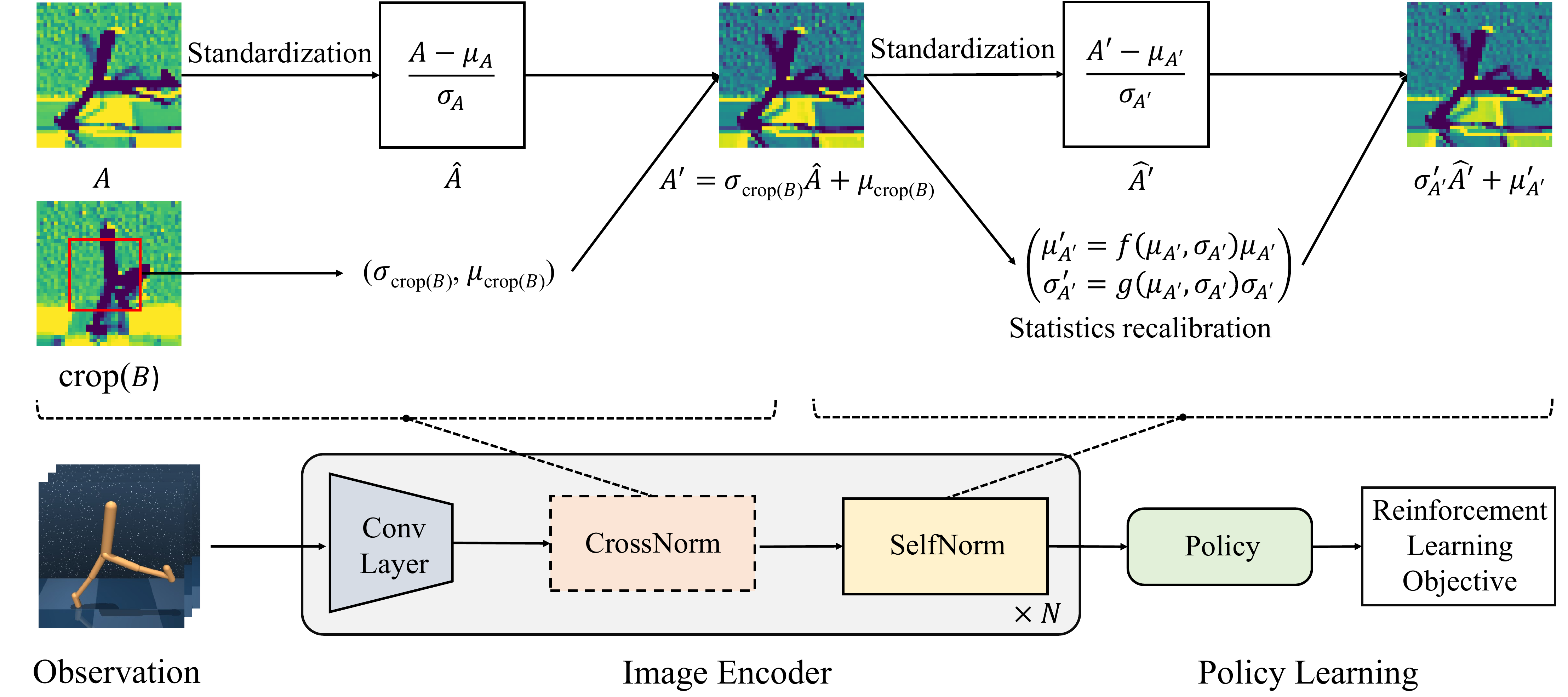}
    \caption{The pipeline of our method. CrossNorm is positioned after the convolutional layer and is followed by SelfNorm. Each CrossNorm layer is randomly activated during training and becomes inactive during testing. Instead, SelfNorm is adopted during training and remains functional during testing. Our method notably does not introduce new learning objective or utilize out-of-domain data.}
    \label{fig:main}
\end{figure}

The primary challenge in visual  RL generalization stems from distribution shifts in observations. This issue is particularly prominent due to the diverse and dynamic nature of environments in RL tasks. Recognizing the proven effectiveness of CrossNorm and SelfNorm in bolstering generalization under distribution shift in computer vision tasks, we explore the possibilities of these normalization techniques in visual RL. By integrating CrossNorm and SelfNorm, we aim to enhance the generalization capability of visual RL, fostering the learning of more robust and generalizable representations.

Although computer vision tasks and visual RL tasks both involve the representation learning of visual input, their respective data distributions can be quite different.  While CrossNorm is inspired by the observation that computer vision datasets are typically rich and diverse, stemming from a variety of sources, visual RL generally involves training the agent within a single task and environment. This situation results in a notably limited data distribution. In other words, the difference between the mean and standard deviation of channel $\mathcal{A}$ and channel $\mathcal{B}$ tends to be small, thus diminishing the effect of the CrossNorm.
Hence, it becomes crucial to further diversify and expand the data distribution.
To achieve this, we utilize random cropping during the computation of the channel's mean $\mu$ and standard deviation $\sigma$, as illustrated in Equation \ref{eq:crop}. 
This technique can result in a wider distribution of the mean and the standard deviation values, further contributing to its ability to adapt to various data distributions.
\begin{equation}\label{eq:crop}
\mathcal{A'} =\sigma_{\mathrm{crop}\mathcal{(B)}}\frac{\mathcal{A}-\mu_\mathcal{A}}{\sigma_\mathcal{A}}+\mu_{\mathrm{crop}\mathcal{(B)}}, \qquad \mathcal{B'} =\sigma_{\mathrm{crop}\mathcal{(A)}}\frac{\mathcal{B}-\mu_\mathcal{B}}{\sigma_\mathcal{B}}+\mu_{\mathrm{crop}\mathcal{(A)}},
\end{equation}
We present the pipeline of our proposed method in Figure~\ref{fig:main}, where our core contribution is the proposal of a plug-and-play module that is equipped with cropped CrossNorm and SelfNorm. Notably, we arrange CrossNorm immediately after the convolution layer, followed by SelfNorm. This sequence is designed to optimally leverage the effects of these two operations, with CrossNorm augmenting the feature diversity before SelfNorm performs intra-instance normalization. Taking into account their characteristics, CrossNorm is activated solely during the training phase, whereas SelfNorm is utilized during the training phase and remains functional during the testing phase.

During each forward pass in the training process, a predetermined number of CrossNorm layers are randomly activated. For these activated layers, each instance in the mini-batch has its $\mu$ and $\sigma$ values for every channel swapped with those of the same channels of another randomly chosen instance. The remaining CrossNorm layers stay inactive during this process. Generally, how many CrossNorm layers can be activated strongly depends on how many hidden layers the encoder of the base algorithm has. We allow a dynamic utilization of the CrossNorm layers because unlike supervised learning, where the model usually has a strong supervised signal and various methods can be applied to learn task-relevant representations, visual RL is lack of sufficient supervised signals. It is thus difficult for it to effectively capture important knowledge from the pixels. As a result, the training process in visual RL is often more fragile and susceptible to disruptions.
By selecting an appropriate number of active CrossNorm layers during the training process, we can effectively manage the learning difficulty, ensuring more stable training dynamics in the learning process.

The role of CrossNorm can be seen as a form of data augmentation. However, unlike traditional data augmentation methods that have been used in visual RL, CrossNorm operates directly on the feature maps rather than the raw observations. This distinction allows CrossNorm to facilitate more diverse alterations. On the other hand, similar to traditional data augmentation methods, CrossNorm improves generalization at the cost of sample efficiency, while SelfNorm aims to offset this trade-off, thereby ensuring a more stable learning process.

Importantly, our method does not introduce new learning objectives or require any out-of-domain data or prior knowledge. This makes it a self-contained and flexible approach to generalization.
Moreover, our method is not only compatible with standard RL algorithms but can also be seamlessly integrated with other techniques aimed at enhancing the generalization of visual RL, and can further improve the robustness of these methods.
This versatility further underscores the generality of our approach.

\section{Experiments}

Our experiments are aimed to investigate the following questions: (1) Does our method enhance the generalization capabilities of vanilla visual RL methods and to what extent does it impact the training performance? (2) Is our proposed method 
general enough to be integrated with existing generalization methods in visual RL to further enhance their capability?

\subsection{Generalization on CARLA Autonomous Driving Tasks}
\subsubsection{Experimental setup}
To assess our method in realistic scenarios and better gauge its effectiveness and generalization capabilities, We evaluate the performance of our method in the CARLA autonomous driving simulator, which offers realistic observations and complex driving scenarios.

We build our method upon DrQ-v2 \cite{yarats2021drqv2} and compare the generalization ability of DrQ-v2+CNSN with state-of-the-art methods and strong baselines:
\textbf{DrQ-v2} \cite{yarats2021drqv2}: our base visual RL algorithm, which is the prior state-of-the-art model-free visual RL algorithm in terms of sample efficiency. It demonstrates superior performance on a variety of tasks while maintaining high sample efficiency, making it a suitable foundation for our research in developing more generalizable visual RL methods.
\textbf{SVEA} \cite{svea}: the previous state-of-the-art data augmentation based method for generalization, which achieves improved performance by reducing Q-variance through the use of an auxiliary loss. 

Our experimental setting in CARLA is adapted from ~\cite{dbc}. To be specific, We utilize three cameras on the vehicle's roof, each providing a 60-degree field of view.  To train the agent and measure the final performance of the different methods, the reward function is defined as Equation \ref{eq:carla-reward}:
\begin{equation}
r_t = \mathbf{v}_{\mathrm{ego}}^{\top} \hat{\mathbf{u}}_{\mathrm{highway}} \cdot \Delta t - \lambda_i \cdot \mathrm{impulse} - \lambda_s \cdot |\mathrm{steer}|,
\label{eq:carla-reward}
\end{equation}
where $\mathbf{v}_{\mathrm{ego}}$ is the velocity vector of the ego vehicle, projected onto the highway's unit vector $\hat{\mathbf{u}}_{\mathrm{highway}}$, and multiplied by the time discretization $\Delta t=0.05$ to measure highway progression in meters. Collisions result in impulses, measured in Newton-seconds. A steering penalty is also applied, with $\mathrm{steer} \in[-1,1]$. The weights used in the reward function are $\lambda_i=10^{-4}$ and $\lambda_s=1$.

\begin{figure}[t]
    \centering
    \includegraphics[width=0.9\linewidth]{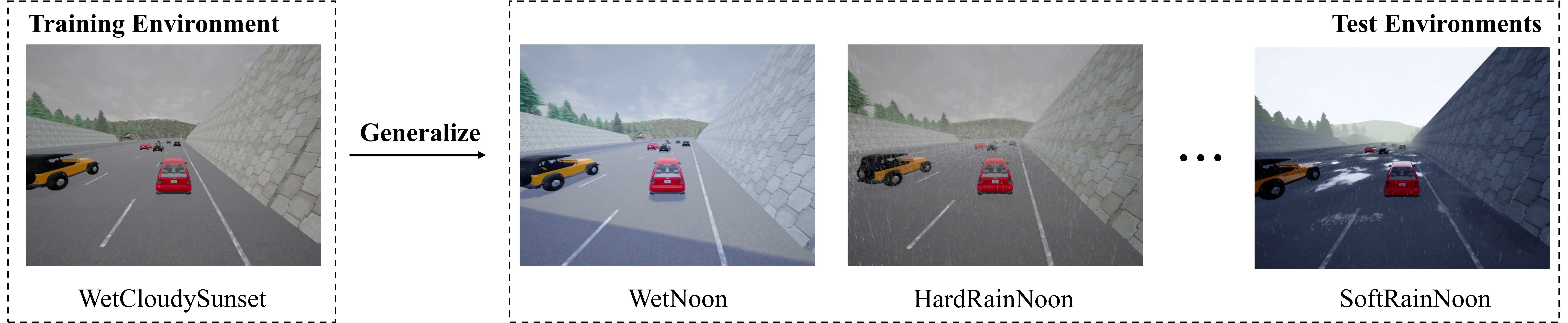}
    \caption{In CARLA autonomous driving simulator, agents are trained under one fixed weather condition. These agents are then expected to generalize to unseen weather conditions in a zero-shot manner. These weather conditions vary in aspects such as lighting, humidity, and other factors, leading to differences not only in visual observation but also in the dynamics of the environment.}
    \label{fig:carla_weather}
\vspace{-0.15in}
\end{figure}

Due to the fact that the encoder in DrQ-v2 has four hidden layers, the maximum number of activated CrossNorm modules in DrQ-v2+CNSN is four. In CARLA experiments, all four CrossNorm layers are activated during the training phase. All agents are trained under one fixed weather condition for 200,000 environment steps. Their performance is then assessed across various other weather conditions within the same map and task, as shown in Figure \ref{fig:carla_weather}. Moreover, it's worth noting that not only the visual observations change with different weather conditions, but also the dynamics of POMDP might vary due to factors like rain.

Since we employ DrQ-v2 as our base visual RL algorithm and baseline method, we also adapt and reimplement SVEA using the DrQ-v2 structure to ensure a fair comparison. We then train the two variations of SVEA on CARLA, one applying random convolution as data augmentation and the other employing random overlay with images from Places365 dataset~\cite{zhou2017places}, respectively.

\subsubsection{Generalization performance}
The generalization performance results are shown in Table \ref{carla}. The results indicate that DrQ-v2 cannot adapt to new weather with different lighting, humidity, \textit{etc}. However, by combing it with CNSN, DrQ-v2+CNSN is enough to generalize well on most of the unseen complicated scenes without a performance drop. Notably, DrQ-v2+CNSN significantly improves the test average performance from DrQ-v2's 14\% of the training performance to \textbf{97\%} of the training performance.

\begin{table}[!htb]
  \caption{\textbf{CARLA generalization results.} Training and testing performance (episode return) of methods trained in 
 one fixed weather and evaluated on other 6 unseen weather conditions. We separately conduct training under two distinct weather conditions: WetCloudySunset (WCS) and HardRainNoon (HRN).
 SVEA(conv) refers to the variant of SVEA that utilizes random convolution for data augmentation, while SVEA(overlay) denotes the variant that employs random overlay for data augmentation. For a fair comparison, we have reimplemented these two versions of SVEA using DrQ-v2.
 The results presented are final performance averaged over 5 random seeds, with each seed corresponding to 50 evaluation episodes for each weather condition.}
  \label{carla}
  \centering

\resizebox{\textwidth}{!}{%
\begin{tabular}{ccccc||cccc}
\toprule
Method & \multicolumn{2}{c}{DrQ-v2} & \multicolumn{2}{c||}{DrQ-v2+CNSN} & \multicolumn{2}{c}{SVEA(conv)} & \multicolumn{2}{c}{SVEA(overlay)} \\
\cmidrule(lr){2-3} \cmidrule(lr){4-5} \cmidrule(lr){6-7} \cmidrule(lr){8-9}
Training Weather& WCS & HRN & WCS & HRN & WCS & HRN & WCS & HRN \\ 
\midrule
Training &$249$\scriptsize$\pm 23$ & $249$\scriptsize$\pm 34$  & $225$\scriptsize$\pm 11$ & $225$\scriptsize$\pm 14$  &$221$\scriptsize$\pm 25$&$243$\scriptsize$\pm 28$  & $173$\scriptsize$\pm 87$ &$204$\scriptsize$\pm 11$ \\ 
\midrule
WetCloudySunset &$249$\scriptsize$\pm 23$ & $118$\scriptsize$\pm 43$&$225$\scriptsize$\pm 11$ &$\bm{211}$\scriptsize$\bm{\pm 9}$  &$221$\scriptsize$\pm 25$ &$184$\scriptsize$\pm 18$ &$173$\scriptsize$\pm 87$ &$30$\scriptsize$\pm 21$ \\ 
MidRainSunset         &  $184$\scriptsize$\pm 18$    & $-2$\scriptsize$\pm 11$
        & $\bm{233}$\scriptsize$\bm{\pm 32}$&$\bm{208}$\scriptsize$\bm{\pm 11}$
        & $184$\scriptsize$\pm 44$&$59$\scriptsize$\pm 91$
        & $160$\scriptsize$\pm 24$&$68$\scriptsize$\pm 22$\\ 
HardRainSunset 
        &  $36$\scriptsize$\pm 26$ &  $-3$\scriptsize$\pm 10$  &$\bm{230}$\scriptsize$\bm{\pm 21}$  &$\bm{221}$\scriptsize$\bm{\pm 16}$
        & $169$\scriptsize$\pm 41$&$79$\scriptsize$\pm 93$
        &$148$\scriptsize$\pm 31$&$87$\scriptsize$\pm 18$\\ 
WetNoon         
&$2$\scriptsize$\pm 6$& $5$\scriptsize$\pm 4$&$\bm{210}$\scriptsize$\bm{\pm 9}$&  $\bm{173}$\scriptsize$\bm{\pm 43}$
        & $82$\scriptsize$\pm 85$& $51$\scriptsize$\pm 53$&$1$\scriptsize$\pm 6$&$-1$\scriptsize$\pm 2$\\ 
SoftRainNoon
        & $-2$\scriptsize$\pm 7$&$-6$\scriptsize$\pm 8$
        &$\bm{232}$\scriptsize$\bm{\pm 40}$ &$\bm{205}$\scriptsize$\bm{\pm 19}$
        & $101$\scriptsize$\pm 90$ &$59$\scriptsize$\pm 69$
        &$57$\scriptsize$\pm 50$&$14$\scriptsize$\pm 26$\\ 
MidRainyNoon
        &  $89$\scriptsize$\pm 38$ &$-3$\scriptsize$\pm 8$
        &  $\bm{237}$\scriptsize$\bm{\pm 27}$  & $\bm{215}$\scriptsize$\bm{\pm 17}$
        & $190$\scriptsize$\pm 38$&$69$\scriptsize$\pm 95$
        & $143$\scriptsize$\pm 29$&$166$\scriptsize$\pm 36$\\ 
HardRainNoon        
&  $145$\scriptsize$\pm 20$ & $249$\scriptsize$\pm 34$ 
&  $\bm{237}$\scriptsize$\bm{\pm 25}$  & $225$\scriptsize$\pm 14$ 
&$190$\scriptsize$\pm 36$ &  $243$\scriptsize$\pm 28$
&$146$\scriptsize$\pm 25$ & $204$\scriptsize$\pm 11$ \\ 
\midrule
Average test return         
&$54$\scriptsize$\pm 56$& $18$\scriptsize$\pm 49$
        & $\bm{230}$\scriptsize$\bm{\pm 29}$ & $\bm{206}$\scriptsize$\bm{\pm 27}$
        &$153$\scriptsize$\pm 75$ & $81$\scriptsize$\pm 88$
        &$109$\scriptsize$\pm 67$&$61$\scriptsize$\pm 61$\\
\bottomrule
\end{tabular}%
}

\end{table}

Moreover, it can be seen that both variants of SVEA, using random convolution and random overlay respectively, exhibit a significant performance drop in unseen weather conditions. For example, SVEA(conv) trained under HardRainNoon achieves an average return of $53$ when tested under WetNoon, while DrQ-v2+CNSN attains an average performance of \textbf{173}, despite the fact that DrQ-v2+CNSN has lower training performance than SVEA(conv). 
The primary reason for the significant performance drop of SVEA is that the two data augmentation techniques it employs do not align well with the test environments. Consequently, these augmentations do not provide sufficient generalization capability for unseen weather conditions, which ultimately limits SVEA's robustness in these scenarios. This finding underscores the necessity for more adaptable and versatile visual RL techniques that can effectively cope with the dynamic and intricate nature of real-world environments. Instead, our method does not rely on any task-specific data augmentation or prior knowledge, and can lead to more robust performance in a wide range of real-world scenarios.

\subsection{Generalization on DMControl Generalization Benchmark}
\subsubsection{Experimental setup}
\begin{figure}[h]
\centering
    \begin{subfigure}[b]{0.115\textwidth}
            \includegraphics[width=\textwidth]{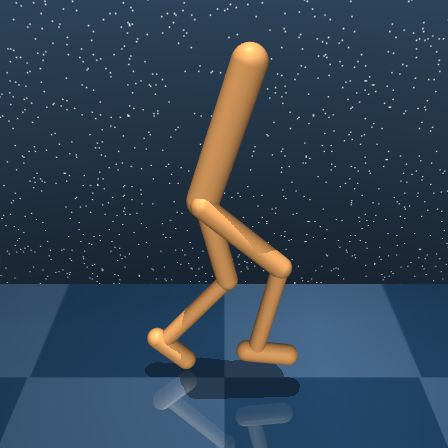}
            \caption{Training}
    \end{subfigure}
    \begin{subfigure}[b]{0.23\textwidth}
            \includegraphics[width=\textwidth]{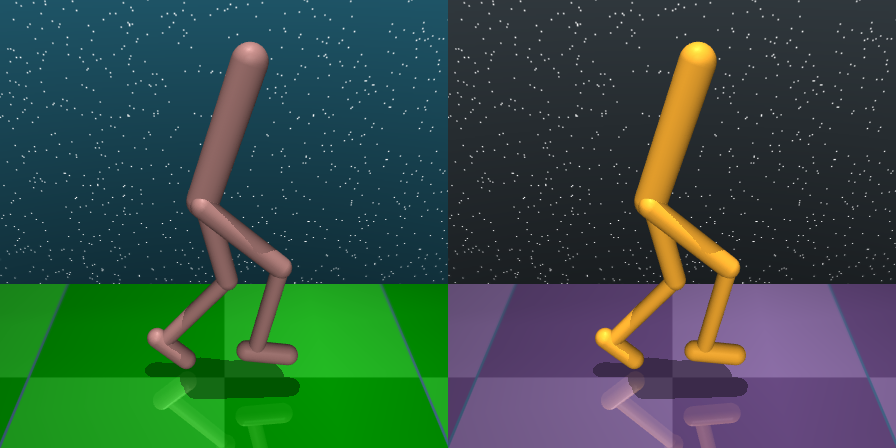}
            \caption{``\emph{Color hard}'' }
    \end{subfigure}
    \begin{subfigure}[b]{0.23\textwidth}
            \includegraphics[width=\textwidth]{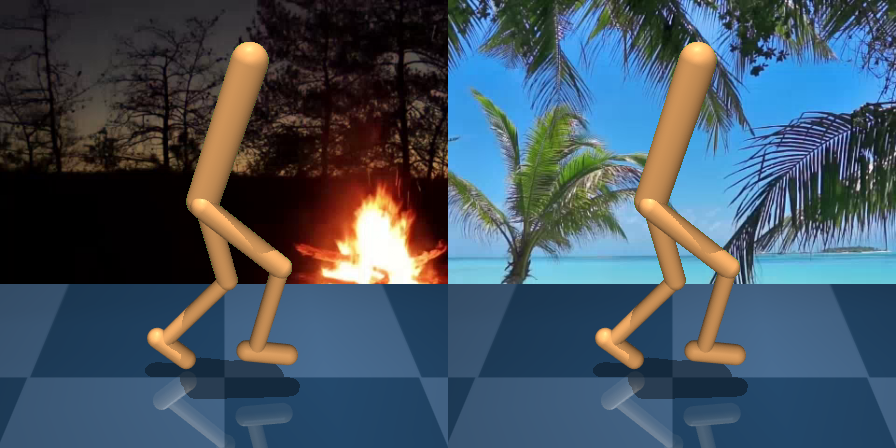}
            \caption{``\emph{Video easy}''}
    \end{subfigure}
    \begin{subfigure}[b]{0.23\textwidth}
            \includegraphics[width=\textwidth]{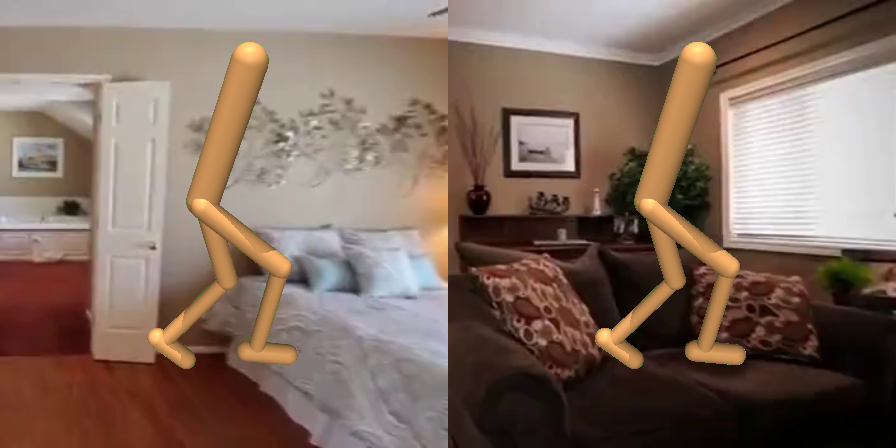}
            \caption{``\emph{Video hard}'' }
    \end{subfigure}
    \caption{Examples of training and testing environments in DMC-GB.}
    \label{fig:dmc}
\end{figure}

We also assess our method on the DeepMind Control Generalization Benchmark (DMC-GB)~\cite{hansen2021softda}, a well-established benchmark for evaluating the generalization capabilities of visual RL algorithms, based on DeepMind Control Suite~\cite{tassa2018deepmind}. In DMC-GB, agents are trained in standard DeepMind Control environments and subsequently evaluated in visually disturbed environments. These disturbances include changes in color (\emph{color hard}) and the replacement of backgrounds with moving videos (\emph{video easy}, \emph{video hard}), as shown in Figure~\ref{fig:dmc}.

For the easy tasks in DeepMind Control Suites, we utilize DrQ as our base visual reinforcement learning algorithm. For medium tasks that DrQ struggles to solve, we employ DrQ-v2 due to its capability to address complex locomotion tasks using pixel observations, providing a more effective solution for these more challenging tasks. To ensure a fair comparison, we have re-implemented SVEA using DrQ-v2 as its base algorithm for medium tasks, considering that the original SVEA was implemented based on DrQ. Our experimental setting mainly follows that of SVEA~\cite{svea}. For the easy tasks, all agents were trained for 500,000 steps in  the vanilla training environments without visual alteration. Meanwhile, for the medium tasks, the training process is extended to 1,500,000 steps for all methods. Note that DrQ contains 11 hidden layers in its encoder while DrQ-v2 only has 4. Across our experiments, we randomly activate 5 out of 11 CrossNorm layers in DrQ-CNSN during the training phase and activate all 4 CrossNorm layers for DrQ-v2.

\subsubsection{Generalization performance}
\begin{table}[]
  \caption{\textbf{DMC-GB generalization results.} Performance on \emph{video easy} and \emph{video hard} testing environments. SVEA refers to the implementation of SVEA that utilizes random overlay as data augmentation method. All the results are averaged over 5 random seeds. \emph{color hard} results can be found in \textbf{Appendix \ref{sec:additional}}.}
  \label{dmcgb}
\begin{center}
    \resizebox{\textwidth}{!}{
    \begin{tabular}{c||c c||c c || c c}
      \toprule
      Easy tasks-\emph{video easy} & DrQ  &  +CNSN  & SVEA& +CNSN & RAD & SODA\\
      \midrule
      Walker Walk 
      & $682$\scriptsize$\pm 89$ 
      & $\bm{792}$\scriptsize$\pm \bm{67}$  
      & $819$\scriptsize$\pm 71$ 
      & $\bm{842}$\scriptsize$\pm \bm{58}$
      & $606$\scriptsize$\pm 63$
      & $635$\scriptsize$\pm 48$ \\
      Walker Stand 
      &  $873$\scriptsize$\pm 83$  
      & $\bm{957}$\scriptsize$\pm \bm{12}$ 
      & $961$\scriptsize$\pm 8$ 
      & $\bm{967}$\scriptsize$\pm \bm{6}$ 
      & $745$\scriptsize$\pm 146$
      & $903$\scriptsize$\pm 56$  \\
      Cartpole Swingup  
      & $485$\scriptsize$\pm 105$ 
      & $\bm{498}$\scriptsize$\pm \bm{26}$ 
      & $\bm{782}$\scriptsize$\bm{\pm 27}$ 
      & $752$\scriptsize$\pm 26$ 
      & $373$\scriptsize$\pm 72$
      & $474$\scriptsize$\pm 143$ \\
      Ball in cup Catch 
      & $318$\scriptsize$\pm 157$  
      & $\bm{584}$\scriptsize$\pm \bm{83}$ 
      & $871$\scriptsize$\pm 106$ 
      & $\bm{913}$\scriptsize$\pm \bm{45}$
      & $481$\scriptsize$\pm 26$
      & $539$\scriptsize$\pm 111$ \\
\toprule
      Medium tasks-\emph{video easy} & DrQ-v2 &  +CNSN & SVEA& +CNSN & RAD & SODA \\
      \midrule
      Cheetah Run 
      & $42$\scriptsize$\pm 19$ 
      & $\bm{274}$\scriptsize$\pm \bm{35}$
      & $\bm{408}$\scriptsize$\bm{\pm 78}$
      & $\bm{404}$\scriptsize$\bm{\pm 29}$
      & - 
      & - \\
      Walker Run  
      & $124$\scriptsize$\pm 31$ 
      & $\bm{452}$\scriptsize$\pm \bm{22}$ 
      & $\bm{611}$\scriptsize$\bm{\pm 20}$ 
      & $\bm{609}$\scriptsize$\bm{\pm 18}$ 
      & - 
      & - \\
  \bottomrule
\end{tabular}}
\end{center}
\begin{center}
\resizebox{\textwidth}{!}{
    \begin{tabular}{c||c c||c c || c c}
      \toprule
      Easy tasks-\emph{video hard} & DrQ &  +CNSN   & SVEA& +CNSN & RAD & SODA\\
      \midrule
      Walker Walk 
      & $104$\scriptsize$\pm 22$ 
      & $\bm{166}$\scriptsize$\pm \bm{28}$ 
      & $377$\scriptsize$\pm 93$ 
      & $\bm{480}$\scriptsize$\pm \bm{46}$  
      & $80$\scriptsize$\pm 10$
      & $312$\scriptsize$\pm 32$\\
      Walker Stand 
      & $289$\scriptsize$\pm 49$ 
      & $\bm{492}$\scriptsize$\pm \bm{62}$ 
      & $834$\scriptsize$\pm 46$ 
      & $\bm{871}$\scriptsize$\pm \bm{23}$  
      & $229$\scriptsize$\pm 45$
      & $736$\scriptsize$\pm 132$ \\
      Cartpole Swingup 
      & $138$\scriptsize$\pm 9$ 
      & $\bm{171}$\scriptsize$\pm \bm{13}$ 
      & $393$\scriptsize$\pm 45$ 
      & $\bm{417}$\scriptsize$\pm \bm{31}$  
      & $152$\scriptsize$\pm 29$
      & $403$\scriptsize$\pm 17$\\
      Ball in cup Catch 
      & $92$\scriptsize$\pm 23$ 
      & $\bm{199}$\scriptsize$\pm \bm{138}$ 
      & $403$\scriptsize$\pm 174$ 
      & $\bm{691}$\scriptsize$\pm \bm{72}$  
      & $98$\scriptsize$\pm 40$
      & $381$\scriptsize$\pm 163$\\
\toprule
      Medium tasks-\emph{video hard} & DrQ-v2 &  +CNSN & SVEA& +CNSN & RAD & SODA\\
      \midrule
      Cheetah Run 
      &$21$\scriptsize$\pm 5$ 
      &$\bm{49}$\scriptsize$\pm \bm{4}$
      &$68$\scriptsize$\pm 9$
      &$\bm{88}$\scriptsize$\pm \bm{9}$
      & - 
      & - \\
      Walker Run 
      &$24$\scriptsize$\pm 2$
      &$\bm{43}$\scriptsize$\pm \bm{2}$
      &$120$\scriptsize$\pm 8$
      &$\bm{148}$\scriptsize$\pm \bm{8}$ 
      & - 
      & - \\
  \bottomrule
\end{tabular}}
\end{center}

\end{table}

To further assess the effectiveness and flexibility of the CrossNorm and SelfNorm in aiding the generalization ability of the visual RL policies, we build CrossNorm and SelfNorm on top of three base visual RL algorithms, DrQ, DrQ-v2, and SVEA. We activate 5 out of 11 CrossNorm layers for SVEA on easy tasks and all 4 CrossNorm layers for SVEA (like DrQ-v2) on medium tasks. We assess the testing performance of DrQ+CNSN, DrQ-v2+CNSN, and SVEA+CNSN across the following settings: \emph{color hard}, \emph{video easy}, and \emph{video hard}, where color-hard tasks have randomly jittered color, video-easy and video-hard tasks replace the background with the unseen moving videos. Notably, the most challenging one is video-hard, where the reference plane of the ground is also removed. We adopt SVEA with random overlay for all these settings and baselines, since it performs better than SVEA(conv) on \emph{video easy} and \emph{video hard} environments. This enables us to investigate whether our module (CrossNorm and SelfNorm) can further enhance generalization when integrated with strong data augmentation-based approaches. Additionally, for comparison purposes, we present the results of two other generalization methods in visual RL, namely RAD~\cite{rad} and SODA~\cite{hansen2021softda}.  As illustrated in Table \ref{dmcgb}, 
incorporating CrossNorm and SelfNorm significantly improves test performance in most of the testing environments compared to the original methods, while maintaining comparable performance in the remaining situations. 
In particular, when applied to DrQ and DrQ-v2, our method achieves substantial improvements in \emph{video easy} and \emph{video hard} environments, with average performance improvement of \textbf{155\%} and \textbf{80\%}, respectively.
Moreover, when combined with SVEA, our method also achieves notable improvements in most environments.
These results further substantiate the efficacy and adaptability of our proposed method. Our methods also outperform RAD and SODA by a large margin on many tasks.

\subsubsection{Sample efficiency and generalization gap}
\begin{figure}[t]
    \centering
    \includegraphics[width=0.24\linewidth]{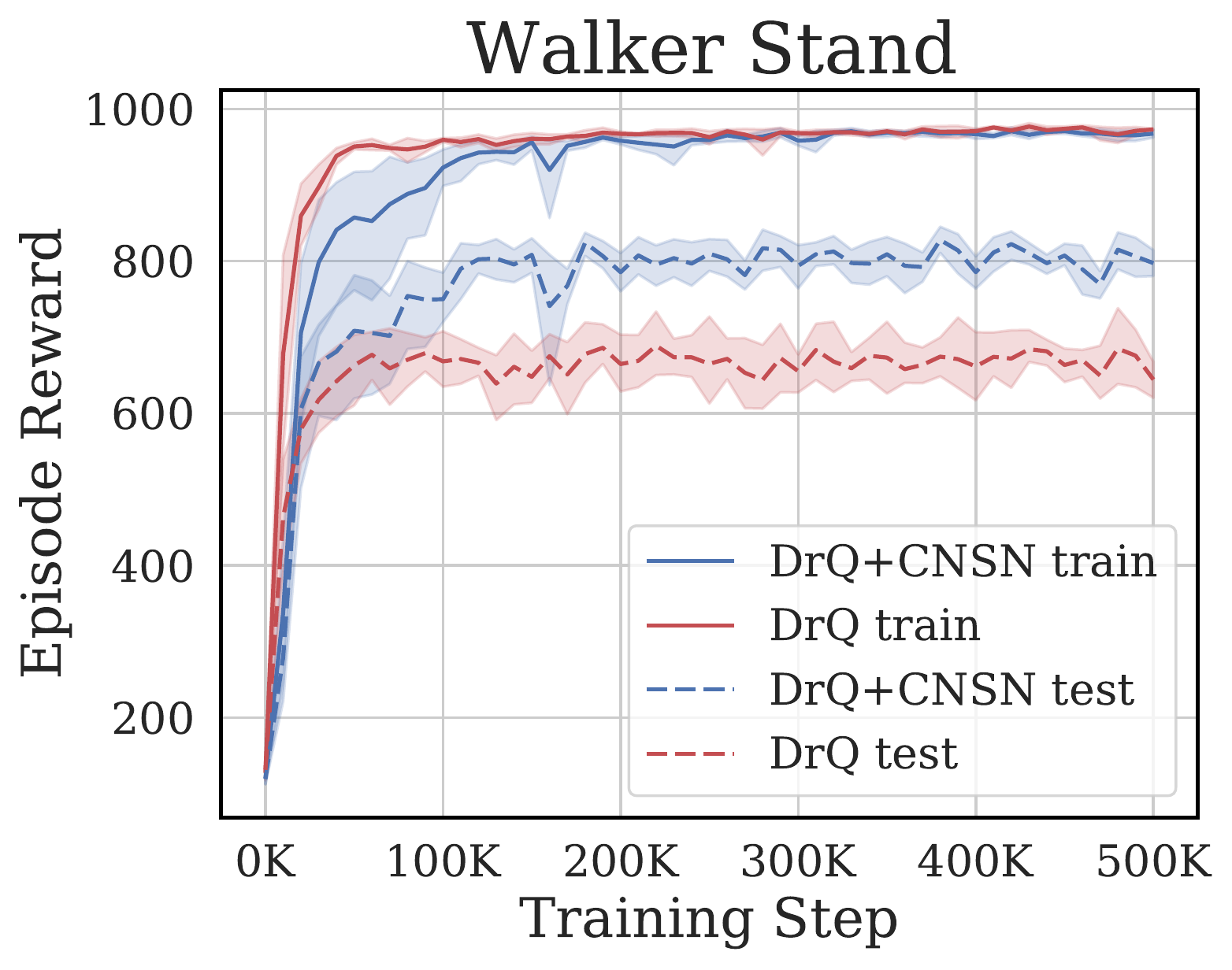}
    \includegraphics[width=0.24\linewidth]{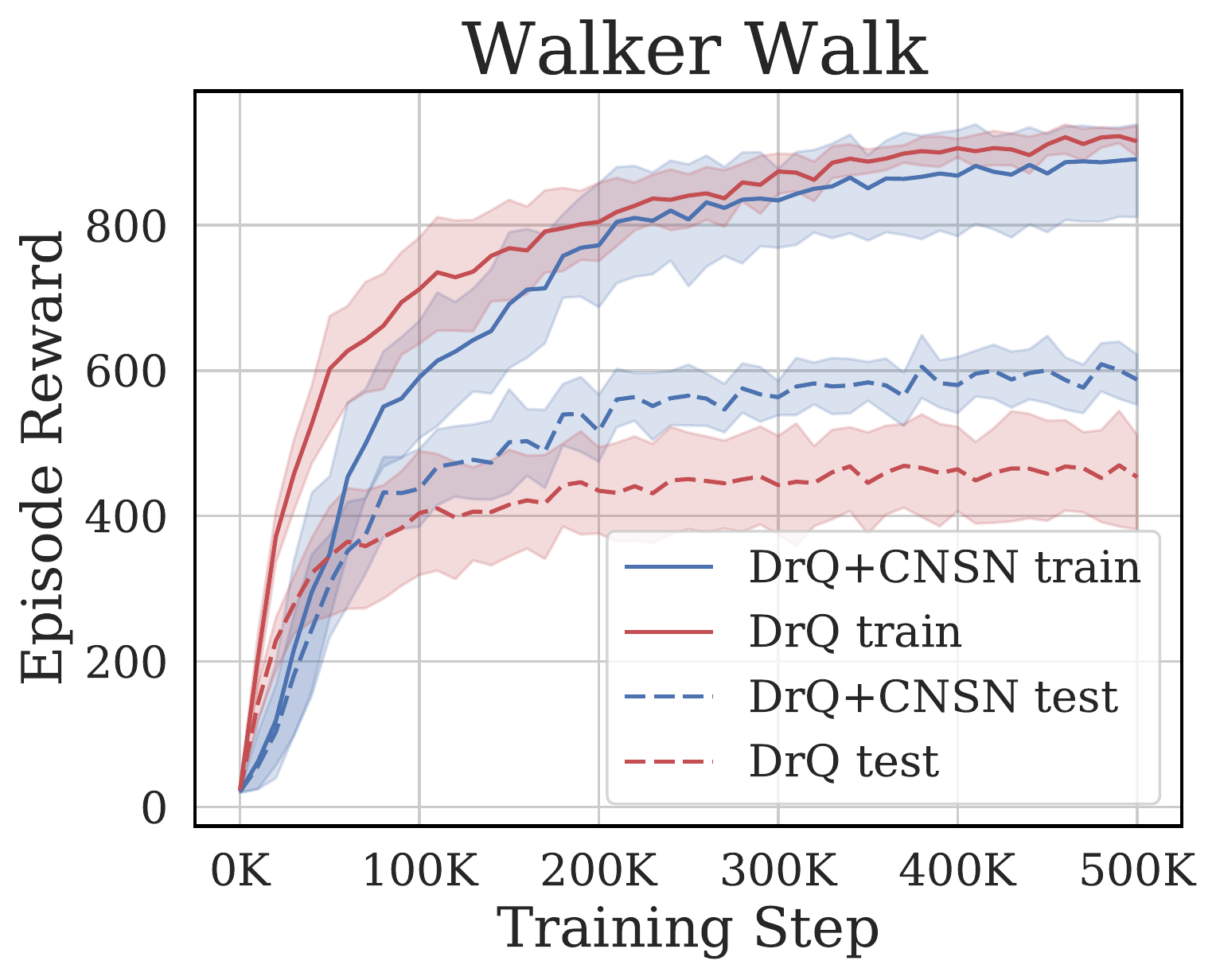} 
    \includegraphics[width=0.24\linewidth]{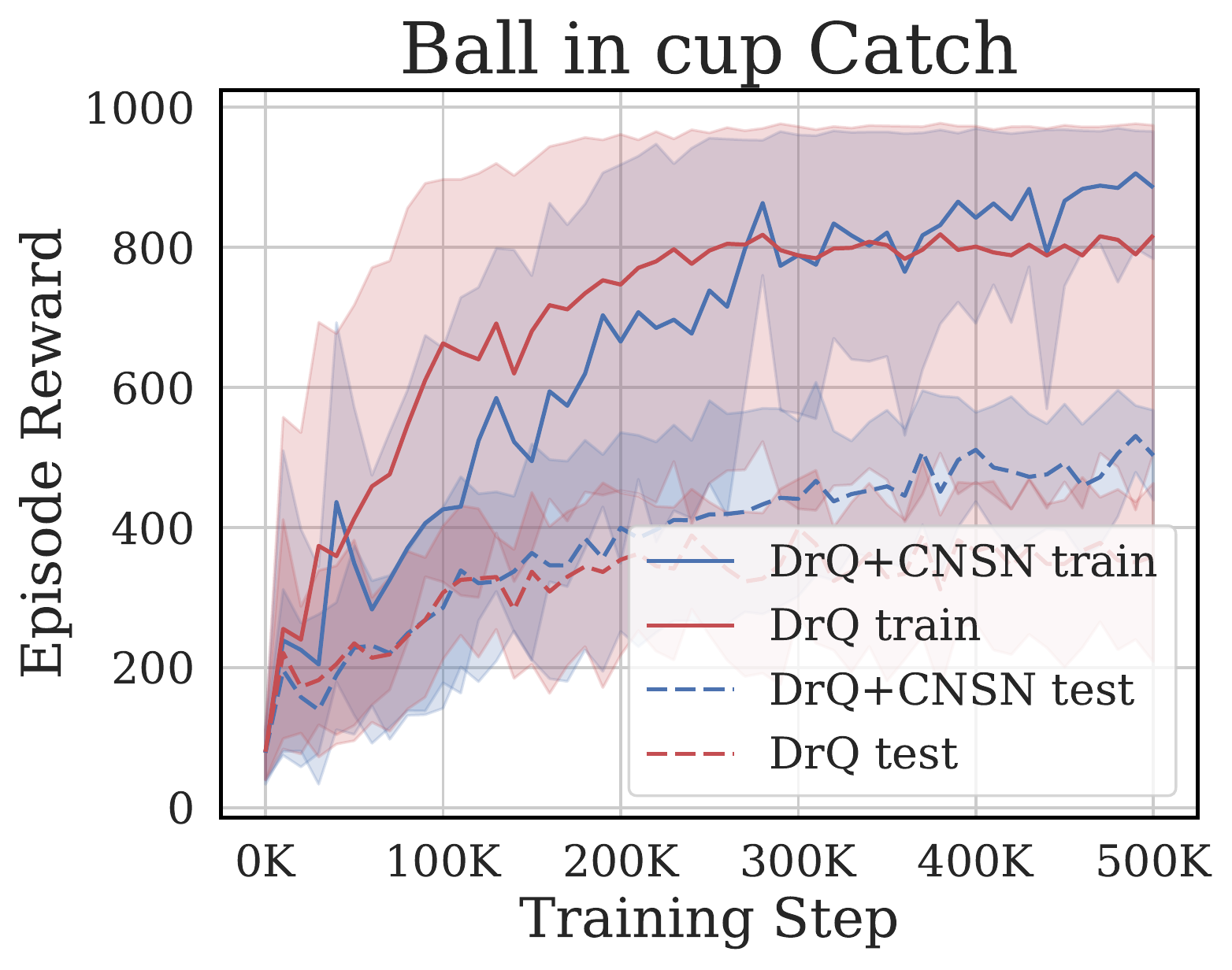} 
    \includegraphics[width=0.24\linewidth]{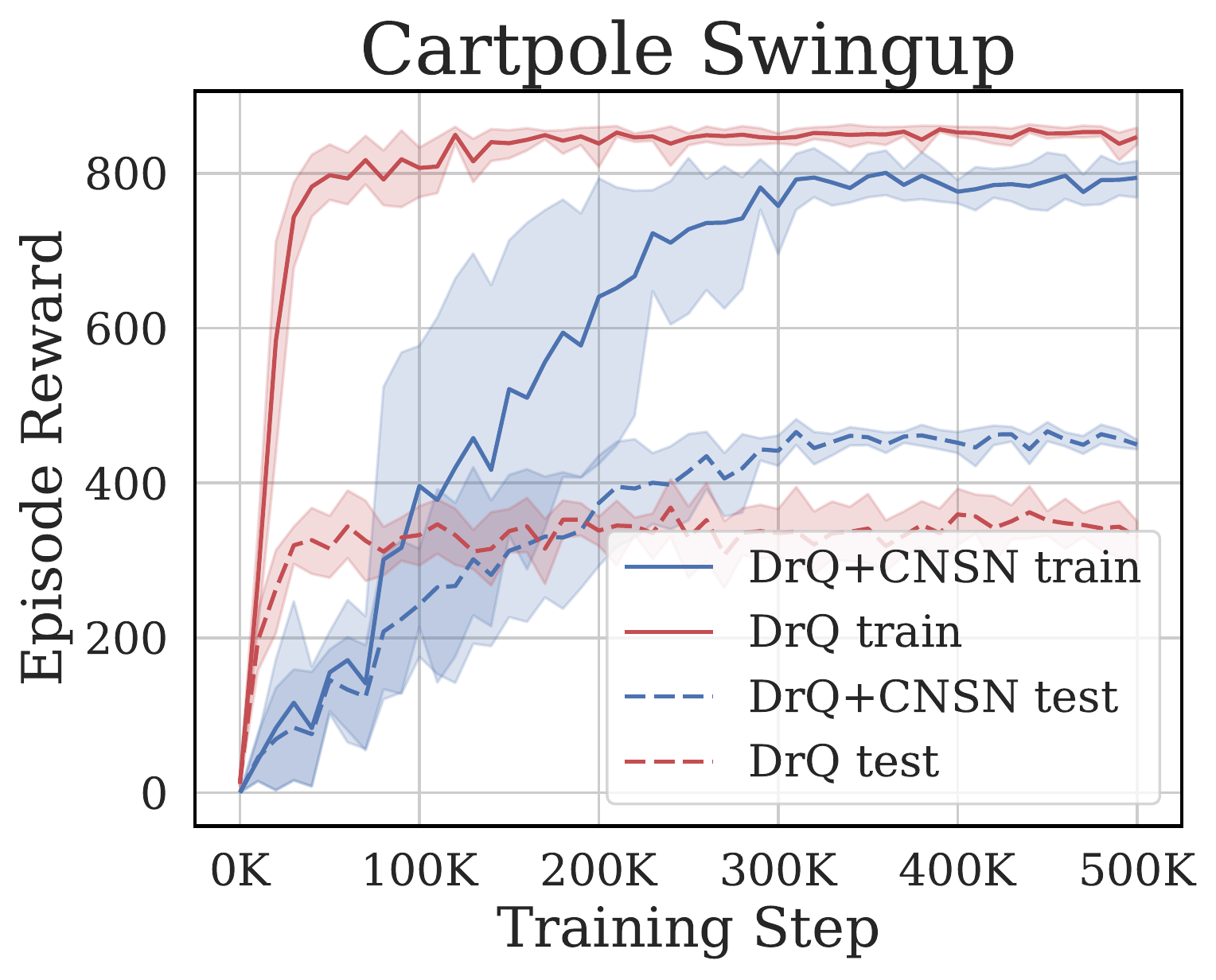}
    \caption{Training and testing performance of DrQ+CNSN against DrQ. The red line is DrQ and blue one corresponds to DrQ+CNSN. The test performance is calculated as the average across the three test settings of DMC-GB, \textit{i.e.}, \emph{color hard}, \emph{video easy}, \emph{video hard}.}
    \label{fig:dmcgb-curve}
\end{figure}

We present the learning curves of DrQ and DrQ+CNSN on four tasks in Figure \ref{fig:dmcgb-curve}. One can find that the generalization gap is significantly reduced by incorporating CrossNorm and SelfNorm. It is worth noting that adopting normalization techniques harms the sample efficiency in the training environments. However, such sacrifice is tolerable since the difference in the training curves on most of the tasks are marginal, while the generalization capability of the agent is largely boosted.

\subsection{Ablation Study}
To validate the essentiality of the design choices incorporated into our method, we perform a series of ablation studies to delve deeper into the understanding of our proposed approach.
\subsubsection{Ablation of CrossNorm and SelfNorm}
Our proposed module is a combination of (cropped) CrossNorm (CN) and SelfNorm (SN). To investigate the individual contributions of CN and SN to generalization capability, we evaluate DrQ+CN and DrQ+SN on several tasks from the DMC-GB and CARLA. This analysis will help us understand the impact of each component on the overall performance of our proposed method. The results are shown in Table \ref{ablation-dmcgb}.

\begin{table}[h]
  \caption{\textbf{Ablation study results.} This table presents the impact of various components on the performance of our method. 
  w/o Crop refers to DrQ+CNSN without using random cropping in CrossNorm. The results of the CARLA benchmark were obtained by training in the WetCloudySunset weather condition and testing in 6 other different weather conditions.}
  \label{ablation-dmcgb}
\begin{center}
\resizebox{\textwidth}{!}{
\begin{tabular}{c|c||c c c c c}
  \toprule
  \multirow{2}{*}{Tasks} & \multirow{2}{*}{Setting} & \multicolumn{5}{c}{Method} \\
  \cmidrule(lr){3-7}
  &  & DrQ  &  +CN &  +SN &  +CNSN &w/o Crop  \\
  \midrule
  \multirow{3}{*}{Walker Walk} & \emph{color hard}& $520$\scriptsize$\pm 91$ &$\bm{823}$\scriptsize$\pm \bm{21}$ &$188$\scriptsize$\pm 34$ & $815$\scriptsize$\pm 65$&$634$\scriptsize$\pm 124$   \\
   & \emph{video easy} & $682$\scriptsize$\pm 89$  &$829$\scriptsize$\pm 60$ &$207$\scriptsize$\pm 39$ & $\bm{842}$\scriptsize$\pm \bm{58}$&$664$\scriptsize$\pm 121$  \\
   & \emph{video hard} & $104$\scriptsize$\pm 22$  &$\bm{196}$\scriptsize$\pm \bm{41}$ &$89$\scriptsize$\pm 24$ & $166$\scriptsize$\pm 28$ &$130$\scriptsize$\pm 35$ \\
  \midrule
  \multirow{3}{*}{Walker Stand} & \emph{color hard} & $770$\scriptsize$\pm 71$ &$\bm{951}$\scriptsize$\pm \bm{27}$ &$525$\scriptsize$\pm 66$ & $942$\scriptsize$\pm 19$  &$841$\scriptsize$\pm 50$ \\
  & \emph{video easy} & $873$\scriptsize$\pm 83$  &$945$\scriptsize$\pm 33$ &$445$\scriptsize$\pm 113$ & $\bm{957}$\scriptsize$\pm \bm{12}$ &$857$\scriptsize$\pm 129$ \\
  & \emph{video hard} & $289$\scriptsize$\pm 49$ &$461$\scriptsize$\pm 81$ &$223$\scriptsize$\pm 22$ & $\bm{492}$\scriptsize$\pm \bm{62}$ &$322$\scriptsize$\pm 46$  \\
  \midrule
  \multirow{3}{*}{Cartpole Swingup} & \emph{color hard} & $586$\scriptsize$\pm 52$  &$\bm{695}$\scriptsize$\pm \bm{38}$ &$187$\scriptsize$\pm 34$ & $679$\scriptsize$\pm 35$ &$560$\scriptsize$\pm 134$ \\
  & \emph{video easy} & $485$\scriptsize$\pm 105$  &$\bm{515}$\scriptsize$\pm \bm{29}$ &$135$\scriptsize$\pm 9$ & $498$\scriptsize$\pm 26$ &$410$\scriptsize$\pm 89$  \\
  & \emph{video hard} & $138$\scriptsize$\pm 9$  &$\bm{183}$\scriptsize$\pm \bm{4}$ &$111$\scriptsize$\pm 22$ &  $171$\scriptsize$\pm 13$ &$155$\scriptsize$\pm 20$  \\
  \midrule
  \multirow{3}{*}{Ball in cup Catch} & \emph{color hard} & $365$\scriptsize$\pm 210$  &$885$\scriptsize$\pm 73$ &$174$\scriptsize$\pm 6$ & $\bm{894}$\scriptsize$\bm{\pm 78}$&$463$\scriptsize$\pm 89$  \\
  & \emph{video easy} & $318$\scriptsize$\pm 157$ &$\bm{599}$\scriptsize$\pm \bm{29}$ &$161$\scriptsize$\pm 33$ & $584$\scriptsize$\pm 83$ &$391$\scriptsize$\pm 116$  \\
  & \emph{video hard} & $92$\scriptsize$\pm 23$  &$146$\scriptsize$\pm 54$ &$75$\scriptsize$\pm 44$ & $\bm{199}$\scriptsize$\pm \bm{138}$&$104$\scriptsize$\pm 35$  \\
    \midrule
  \multirow{1}{*}{CARLA} & unseen weather 
  &$54$\scriptsize$\pm 56$
  &$183$\scriptsize$\pm 91$
  &$71$\scriptsize$\pm 70$
  &$\bm{230}$\scriptsize$\bm{\pm 29}$
  &$185$\scriptsize$\pm 94$\\
  \bottomrule
\end{tabular}}
\end{center}

\end{table}

As previously mentioned, while computer vision datasets often originate from diverse sources, the training of visual RL agents typically occurs within a single task and environment, leading to a relatively narrow data distribution compared to that of computer vision data. Therefore, it's understandable that using SelfNorm alone aids computer vision tasks but could reduce the robustness of visual RL. The $\mu$ and $\sigma$ of the feature maps tend to be relatively stable, causing SelfNorm to overfit, which ultimately leads to a decrease in generalization performance.

It seems that using CrossNorm alone upon DrQ sometimes results in comparable test performance against DrQ+CNSN. However, in more complex autonomous driving scenarios, we observe that relying solely on CrossNorm does not yield performance as good as using both CrossNorm and SelfNorm.
The results suggest that SelfNorm may only be effective in visual RL tasks with the existence of CrossNorm. Furthermore, the empirical results in CARLA scenarios also validate that. It is interesting to note here that it seems that for complex real-world applications, it is beneficial to combine the above two normalization techniques. 

\subsubsection{Ablation on the random cropping of CrossNorm}
We also investigate how random cropping of CrossNorm (Equation \ref{eq:crop}) helps the generalization in DMC-GB tasks, as shown in Table ~\ref{ablation-dmcgb}. The results show that the inclusion of random cropping when calculating $\mu$ and $\sigma$ in CrossNorm significantly improves generalization performance compared to cases without cropping.

\section{Conclusion}
In this paper, we explore the potential benefits of normalization techniques on the generalization capabilities of visual RL and propose a novel normalization module containing CrossNorm and SelfNorm for generalizable RL. By conducting extensive experiments upon different base algorithms across diverse tasks in two generalization benchmarks, DMC-GB and CARLA autonomous driving simulator, we demonstrate that our method is able to enhance generalization capability without the help of out-of-domain data and prior knowledge. These characteristics establish our approach as a self-contained method for achieving generalizable visual RL. Our method can be integrated with any visual RL algorithm, making it a valuable approach for tackling unpredictable environments.

\textbf{Limitation}: The limitations of our method lie in (1) one needs to predetermine the number of activated CrossNorm layers, and may need some experimentation to obtain optimal results; (2) since our method does not employ out-of-domain data, its generalization performance may not surpass those using pre-trained models~\cite{Yuan2022PreTrainedIE}. It is interesting to see whether combining CNSN with pre-trained models can further benefit generalization, which we leave as future work.

{
\small
\bibliographystyle{plain}
\bibliography{main}
}

\clearpage
\appendix

\vspace{7pt}
\section*{\Large \centering Appendix}
\vspace{7pt}

\vspace{20pt}

\section{Additional Results}
In this section, we  provide additional results to further illustrate the findings of our study.
\subsection{Additional DMC-GB generalization results}
\label{sec:additional}
As illustrated in Table~\ref{add-dmcgb}, the results from the \emph{color hard} environments further confirm that the combination of CrossNorm and SelfNorm significantly enhances the generalization performance under \emph{color hard} evaluation setting.
\begin{table}[h]
\caption{\textbf{DMC-GB \emph{color hard} generalization results.} Performance comparison of DrQ and DrQ+CNSN on \emph{color hard} testing environment. All the results are averaged over 5 random seeds.}
\label{add-dmcgb}
\begin{center}
 \begin{tabular}{c||c c}
      \toprule
 Easy tasks-\emph{color hard}& DrQ &  +CNSN \\
      \midrule
      Walker Walk 
      & $520$\scriptsize$\pm 91$
      & $\bm{815}${\scriptsize$\pm \bm{65}$} (+\textbf{56.7\%})   \\
      Walker Stand 
      &  $770$\scriptsize$\pm 71$
      & $\bm{942}${\scriptsize$\pm \bm{19}$} (+\textbf{22.3\%})  \\
      Cartpole Swingup 
      & $586$\scriptsize$\pm 52$ 
      & $\bm{679}${\scriptsize$\pm \bm{35}$} (+\textbf{15.9\%})   \\
      Ball in cup Catch 
      & $365$\scriptsize$\pm 210$ 
      & $\bm{894}${\scriptsize$\pm \bm{78}$} (+\textbf{144.9\%}) \\
    \toprule
     Medium tasks-\emph{color hard} & DrQ-v2 &  +CNSN \\
      \midrule
      Cheetah Run 
      & $144$\scriptsize$\pm 29$ 
      & $\bm{345}${\scriptsize$\pm \bm{57}$} (+\textbf{139.6\%})  \\
      Walker Run 
      & $90$\scriptsize$\pm 21$ 
      & $\bm{429}${\scriptsize$\pm \bm{16}$} (+\textbf{376.7\%})  \\
  \bottomrule
\end{tabular}
\end{center}
\end{table}

\subsection{Generalization Performance Comparison against other Normalization Techniques}

In addition to CrossNorm and SelfNorm, we also investigate two other types of normalization techniques prevalent in deep learning: batch normalization (BN) and spectral normalization (SpecN). We integrate each of these into the image encoder of DrQ separately to assess their potential to enhance the generalization performance. BN layers are positioned after every convolution layer in the image encoder. 
When utilizing SpecN, we follow the conclusion proposed by [15] that using too many SpecN layers can decrease the capacity of networks and be detrimental to learning. Therefore, in our setting, SpecN layers are only placed after the second, third, and fourth convolution layers of the image encoder.
We train these agents on two DMC-GB tasks and evaluate their generalization performance in three settings.

\begin{table}[h]
  \caption{Comparison of generalization performance with different normalization techniques on DMC-GB. The results demonstrate that BN and SpecN do not lead to improvements in generalization performance under distribution shift.}
  \label{BN-SpecN}
  
\begin{center}
\begin{tabular}{c|c||c c c c}
  \toprule
  \multirow{2}{*}{Tasks} & \multirow{2}{*}{Setting} & \multicolumn{4}{c}{Method} \\
  \cmidrule(lr){3-6}
  &  & DrQ  &  +BN &  +SpecN &+CNSN \\
  \midrule
  \multirow{3}{*}{Walker Walk} & \emph{color hard}& $520$\scriptsize$\pm 91$   & $257$\scriptsize$\pm 89$&$525$\scriptsize$\pm 64$&$\bm{815}$\scriptsize$\pm \bm{65}$ \\
   & \emph{video easy} & $682$\scriptsize$\pm 89$ & $479$\scriptsize$\pm 109$&$739$\scriptsize$\pm 19$  &$\bm{842}$\scriptsize$\pm \bm{58}$ \\
   & \emph{video hard} & $104$\scriptsize$\pm 22$ & $57$\scriptsize$\pm 12$&$145$\scriptsize$\pm 20$ &$\bm{166}$\scriptsize$\pm \bm{28}$\\
  \midrule
  \multirow{3}{*}{Cartpole Swingup} & \emph{color hard} & $586$\scriptsize$\pm 52$ &$164$\scriptsize$\pm 48$ &$512$\scriptsize$\pm 107$ &$\bm{679}$\scriptsize$\pm \bm{35}$\\
  & \emph{video easy} & $485$\scriptsize$\pm 105$ & $182$\scriptsize$\pm 66$ &$375$\scriptsize$\pm 14$ & $\bm{498}$\scriptsize$\pm \bm{26}$\\
  & \emph{video hard} & $138$\scriptsize$\pm 9$  & $113$\scriptsize$\pm 13$& $130$\scriptsize$\pm 2$ & $\bm{171}$\scriptsize$\pm \bm{13}$\\
  \bottomrule
\end{tabular}
\end{center}
\end{table}

As shown in Table ~\ref{BN-SpecN}, the results show that both BN and SpecN do not improve the generalization performance. Furthermore, BN leads to a significant decrease in generalization capabilities. This can be attributed to the fact that BN assumes the test data distribution is the same as the training data distribution, which can result in performance degradation when facing distribution shift. Previous literature suggests that SpecN is effective in maintaining a stable learning process for RL, particularly for very deep neural networks. Based on our results, It appears that SpecN does not significantly affect the generalization performance when faced with visual disturbances.

\subsection{Parameter Study}
The number of active CrossNorm layers is a crucial hyper-parameter in our method. In this section, we discuss how the performance of our approach is affected by varying the number of active CrossNorm layers. Our investigation reveals that activating too many CrossNorm layers when combining them with DrQ during training can lead to divergence in the learning process. For example, when applied to the Cartpole Swingup task, activating all 11 CrossNorm layers resulted in divergence across all 5 random seeds, leading to an average reward of only 158 in the training environments. As a result, to ensure generalizability and consistency in our experiments, we choose to activate only 5 out of the 11 CrossNorm layers for the DrQ+CNSN configuration. 
On the other hand, the difference between DrQ and DrQ-v2 lies in their encoder architecture and algorithm. We found that activating all 4 CrossNorm layers for DrQ-v2+CNSN did not result in divergence in our experiments.
This allows us to fully leverage the benefits of CrossNorm in enhancing the generalization performance of DrQ-v2+CNSN. This observation highlights the importance of  activating the appropriate number of CrossNorm layers for each algorithm to ensure stable training. The optimal number of active  CrossNorm layers may indeed vary depending on the specific encoder architecture and algorithm employed.

\section{Training Details}
In this section, we provide our detailed settings in experiments. All experiments were conducted using a single GeForce GTX 3090 GPU and Intel Xeon Silver 4210 CPUs. All code assets used for this project came with MIT licenses. \textbf{Our code} for the CARLA and DMC-GB experiments can be found in the supplementary material.
\subsection{CARLA experiments}
While the majority of the hyper-parameters remain the same as in the original implementation of DrQ-v2, several were modified to better adapt to the CARLA environments. The complete hyper-parameter settings are presented in Table~\ref{table:CARLA_hp}. All agents are trained under a fixed weather condition for 200,000 environment steps, and their performance is then evaluated in six unseen weather conditions within the same map and task.

\begin{table}[htbp]
\caption{\label{table:CARLA_hp} A default set of hyper-parameters used in CARLA experiments.}
\renewcommand{\arraystretch}{1.15}
\centering
\resizebox{0.6\textwidth}{!}{
\begin{tabular}{lc}
\toprule
\multicolumn{2}{c}{DrQ-v2 Hyper-parameters} \\
\midrule
Frame rendering &$84 \times 252 \times 3$ \\
Stacked frames & $3$ \\
Replay buffer capacity & 100,000 \\
Action repeat & $4$ \\
Exploration steps & $100$ \\
$n$-step returns & $3$ \\
Batch size & $512$ \\
Discount $\gamma$ & $0.99$ \\
Optimizer & Adam \\
Learning rate & 1e-4 \\
Agent update frequency & $2$ \\
Critic Q-function soft-update rate $\tau$ & $0.01$ \\
Exploration stddev. clip & $0.3$ \\
Exploration stddev. schedule & $\mathrm{linear}(1.0, 0.1, 100000)$\\ 
\midrule
\multicolumn{2}{c}{CrossNorm Hyper-parameters} \\
\midrule
active CrossNorm  & 4 out of 4\\
\midrule
\multicolumn{2}{c}{SVEA Hyper-parameters} \\
\midrule
SVEA coefficients
 &  $\alpha=0.5,\beta=0.5$\\

\bottomrule
\end{tabular}}
\end{table}

\subsection{DMC-GB experiments}
For the DrQ algorithm, we adopt the exact same hyper-parameters as those used in the implementation of DrQ in the DMC-GB. For DrQ-v2, we adhere to the same hyper-parameters as the original implementation of DrQ-V2 for DMC tasks. The complete hyper-parameter settings are presented in Table ~\ref{table:dcmgb_hp}. For the easy tasks in the DeepMind Control (DMC), we utilize the DrQ algorithm as the base RL algorithm and train the agents for 500,000 environment steps. For the medium tasks, we adopt the DrQ-v2 algorithm as the base RL algorithm for all methods, and the training steps are extended to 1,500,000 steps.
\begin{table}[htbp]
\caption{A default set of hyper-parameters used in DMC-GB experiments.}
\label{table:dcmgb_hp}
\renewcommand{\arraystretch}{1.15}
\centering
\resizebox{0.6\textwidth}{!}{
\begin{tabular}{lc}
\toprule
\multicolumn{2}{c}{DrQ Hyper-parameters} \\
\midrule
Action repeat& 2\\
Discount $\gamma$                                                         & 0.99                                                                              \\
Replay buffer size                                                               & 500,000                                                                           \\
Optimizer   & Adam                                                       \\
Learning rate ($\theta$)                                                         & 1e-3                                                      \\
Learning rate ($\alpha$ of SAC)                                                  & 1e-4                                                                              \\
Batch size                                                                    & 128 \\
\midrule
\multicolumn{2}{c}{DrQ-v2 Hyper-parameters} \\
\midrule
Replay buffer size & 1,000,000 \\
Action repeat & $2$ \\
Exploration steps & $2000$ \\
$n$-step returns & $3$ \\
Batch size & $256$ \\
Discount $\gamma$ & $0.99$ \\
Optimizer & Adam \\
Learning rate & 1e-4 \\
Agent update frequency & $2$ \\
Critic Q-function soft-update rate $\tau$ & $0.01$ \\
Exploration stddev. clip & $0.3$ \\
Exploration stddev. schedule & $\mathrm{linear}(1.0, 0.1, 500000)$\\ 
\midrule
\multicolumn{2}{c}{CrossNorm Hyper-parameters} \\
\midrule
active CrossNorm (w/ DrQ) & 5 out of 11\\
active CrossNorm (w/ DrQ-v2) & 4 out of 4\\
\midrule
\multicolumn{2}{c}{SVEA Hyper-parameters} \\
\midrule
SVEA coefficients
 &  $\alpha=0.5,\beta=0.5$\\

\bottomrule
\end{tabular}}
\end{table}

\end{document}